\documentclass[submission,copyright,creativecommons]{eptcs}
\providecommand{\event}{ICLP 2019} 
\usepackage{breakurl}             
\usepackage{underscore}           

\usepackage{mathptmx}

\usepackage{times}
\usepackage{soul}
\usepackage{url}
\usepackage[utf8]{inputenc}
\usepackage[small]{caption}
\usepackage{graphicx}
\usepackage{amsmath}
\usepackage{booktabs}
\usepackage{algorithm}
\usepackage{algorithmic}
\usepackage[toc,page]{appendix}
\usepackage{color}
\usepackage{multicol}
\usepackage{tabularx,ragged2e}
\usepackage{makecell}
\usepackage{CJKutf8}

\usepackage{listings}
\lstdefinelanguage{clingo}{
  keywordstyle=[1]\usefont{OT1}{cmtt}{m}{n},%
  keywordstyle=[2]\textbf,%
  keywordstyle=[3]\usefont{OT1}{cmtt}{m}{n},
  alsoletter={\#,\&},%
  keywords=[1]{not,from,import,def,if,else,return,while,break,and,or,for,in,del,and,class},%
  keywords=[2]{\#const,\#show,\#minimize,\#base,\#theory,\#count,\#external,\#program,\#script,\#end,\#heuristic,\#edge,\#project,\#show},%
  keywords=[3]{&,&dom,&sum,&diff,&show,&minimize},%
  morecomment=[l]{\#\ },%
  morecomment=[l]{\%\ },%
  commentstyle={\color{darkgray}}%
}
\lstdefinelanguage{gf}{
  keywordstyle=[1]\usefont{OT1}{cmtt}{m}{n},%
  keywordstyle=[2]\textbf,%
  keywordstyle=[3]\usefont{OT1}{cmtt}{m}{n},
  alsoletter={\#,\&},%
  keywords=[1]{concrete,lincat,lin,of,abstract,flags,cat,fun,open,in,oper},%
  morecomment=[l]{\#\ },%
  morecomment=[l]{\%\ },%
  commentstyle={\color{darkgray}}%
}
\newcommand{\ar}{{$\rightarrow$ }}

\title{Natural Language Generation for Non-Expert Users}
\author{
Van Duc Nguyen \quad Tran Cao Son \quad Enrico Pontelli
\institute{New Mexico State University\\ Las Cruces, New Mexico, USA}
\email{vnguyen|tson|epontell@cs.nmsu.edu}
}
\def\titlerunning{Natural Language Generation for Non-Expert Users}
\def\authorrunning{V.D. Nguyen, T. C. Son \& E. Pontelli}
\begin{document}
\maketitle

\begin{abstract}
Motivated by the difficulty in presenting computational results---especially when the results are a collection of atoms in a logical language---to users, who are not proficient in computer programming and/or the logical representation of the results, we propose a system for automatic generation of \emph{natural language descriptions} for applications targeting mainstream users. Differently from many earlier systems with the same aim, the proposed system does not employ templates for the generation task. It assumes that there exist some natural language sentences in the application domain and uses this repository for the natural language description. It does not require, however, a large corpus as it is often required in machine learning approaches. The systems consist of two main components. The first one aims at analyzing the sentences and constructs a Grammatical Framework (GF) for given sentences and is implemented using the Stanford parser and an answer set program. The second component is for sentence construction and relies on GF Library. The paper includes two use cases to demostrate the capability of the system. As the sentence construction is done via GF, the paper includes a use case evaluation showing that the proposed system could also be utilized in addressing a challenge to create an abstract Wikipedia, which is recently discussed in the BlueSky session of the 2018 International Semantic Web Conference.
\end{abstract}

\section{Introduction}

Natural language generation (NLG) has been one of the key topics of research in natural language processing, which was highlighted by the huge body of work on NLG surveyed in \cite{ReiterD00,GattK18}. With the advances of several devices capable of understanding spoken language and conducting conversation with human (e.g., Google Home, Amazon Echo) and the shrinking gap created by the digital devices, it is not difficult to foresee that the market and application areas of NLG systems will continue to grow, especially in applications whose users are non-experts. In such application, a user often asks for certain information and waits for the answer and a NLG module would return the answer in spoken language instead of text such as in question-answering systems\footnote{E.g., the Ergo system: \url{http://coherentknowledge.com}} or recommendation systems\footnote{\url{http://gem.med.yale.edu/ergo/default.htm}}. The NLG system in these two applications uses templates to generate the answers in natural language for the users. A more advanced NLG system in this direction is described in \cite{van2019padl}, which works with ontologies annotated using the Attempto language and can generate a natural language description for workflows created by the systems built in the Phylotastic project\footnote{The Phylotastic project: \url{http://phylotastic.org}}. The applications targeted by these systems are significantly different from NLG systems, whose main purpose is to generate high-quality natural language description of objects or reports, such as those reported in the recent AAAI conference \cite{liu2019table,cyras2019argumentation,arindam2019declarative}.

The present paper is motivated by the need to generate natural language description of computational results to non-expert users such as those developed in the Phylotastic project. In this project, the users are experts in evolutionary biology but are none experts in ontologies and web services. When a user places a request, he/she will receive a workflow consisting of web services, whose inputs and outputs are specified by instances of classes in the ontologies working with web services, as well as the ordering and relationships between the services. To assist the user in understanding the workflow, a natural language description of the workflow is generated. In order to accomplish the task, the NLG system in the Phylotastic project proposes to annotate elements of the ontologies using Attempto, a simple subset of English with precisely defined syntax and semantics.

In this paper, we propose a system that addresses the limitation of the system discussed in the Phylotastic project \cite{van2019padl}. Specifically, we assume that the annotations given in an ontology are natural language sentences. This is a reasonable assumption given that the developers of an ontology are usually those who have intimate knowledge about entities described in the ontology and often have some sort of comments about classes, objects, and instances of the ontology. We then show that the system is very flexible and can be used for the same purpose with new ontologies.

The rest of the paper is organized as follows. Section~\ref{background} briefly reviews the basics of Grammatical Framework (GF)\cite{ranta2004grammatical}. Section~\ref{method} describes the main modules of the system. Section~\ref{usecase} includes two use cases of the system using an available ontologies against in the context of reasoning about ontologies. Specifically, it compares with the system used in the Phylotastic project and an ontology about people. This section also contains a use case that highlights the versatility of the proposed system by addressing a challenge to create an abstract Wikipedia \cite{Vrandecic18}.
Related works are discussed in Section~\ref{relatedwork}. Section~\ref{conclusion} concludes the paper.

\section{Background: Grammatical Framework} 
\label{background}

The Grammatical Framework (GF) \cite{ranta2004grammatical} is a system used for working with grammars. The GF Resource Grammar Library (RGL)\footnote{\url{https://www.grammaticalframework.org/lib/doc/synopsis/index.html}} covering syntax of various languages is the standard library for GF. A GF program has two main parts. The first part is the \emph{Abstract syntax} which defines what meanings can be expressed by a grammar. The abstract syntax defines categories (i.e., types of meaning) and functions (i.e., meaning-building components). An example of an abstract syntax:  

{
\begin{lstlisting}[language=gf,label=gfabstract,caption=Abstract syntax] 
abstract People = {
  flags startcat = Message ;
  cat Message ; People ; Action ; Entity ; 
  fun simple_sent : People -> Action -> Entity -> Message ;
      Bill : People;  Play : Action;   Soccer : Entity ; }

\end{lstlisting}
}

\noindent 
Here, {\tt \small Message}, {\tt \small People}, {\tt \small Action} and {\tt \small Entity} are types of meanings. {\tt \small startcat flag} 
states that {\tt \small Message} is the default start category for parsing and generation. {\tt \small simple\_sent} is a function accepting 3 parameters, of type {\tt \small People}, {\tt \small Action}, {\tt \small Entity}. This function returns a meaning of {\tt \small Message} category. Intuitively, each function in the abstract syntax represents a rule in a grammar. The combination of rules used to construct a meaning type can be seen as a syntax tree. 

The second part is composed of  \emph{one or more concrete syntax specifications.} Each concrete syntax defines the representation of meanings in each output language. For example, to demostrate the idea that one meaning can be represented by different concrete syntaxes, we create two concrete syntaxes for two different languages: English and Italian. To translate a sentence to different languages, we only need to provide the strings representing each word in corresponding languages. The GF libraries will take responsibility to concatenate the provided strings according to the language grammar to create a complete sentence, which is the representations of the meaning, in the targeted language. The corresponding concrete syntaxes that map functions in the 
abstract grammar above to strings in English and in Italian is:

\begin{lstlisting}[language=gf,escapechar=\#,caption=Concrete English and Italian syntaxes,label=gfconcrete]
concrete PeopleEng of People =
open SyntaxEng, ParadigmsEng, ConstructorsEng in {
lincat
  Message = Cl ; People = NP ; Action = V2 ; Entity = NP ;
lin
  simple_sent People Action Entity = mkCl People (mkVP Action Entity) ;
  Bill = mkNP Bill_N; Play = play_V2; Soccer = mkNP soccer_N;
oper
  Bill_N = mkN "Bill" "Bill"; play_V2 = mkV2 "play"; 
  soccer_N = mkN "soccer"; }
  
concrete PeopleIta of People =
open SyntaxIta, ParadigmsIta, ConstructorsIta in {
lincat
  Message = Cl ; People = NP ; Action = V2 ; Entity = NP ;
lin
  simple_sent People Action Entity = mkCl People (mkVP Action Entity) ;
  Bill = mkNP Bill_N; Play = play_V2; Soccer = mkNP soccer_N ;
oper
  Bill_N = mkN "Bill"; play_V2 = mkV2 "giocare"; 
  soccer_N = mkN "calcio" ; }  
\end{lstlisting}

%

%
%
\noindent In these concrete syntaxes, the linearization type definition ({\tt \small lincat}) states that {\tt \small Message}, {\tt \small People}, {\tt \small Action} and {\tt \small Entity} are type {\tt \small Cl} (clause), {\tt \small NP} (noun phrase), {\tt \small V2} (two-place verb), and {\tt \small NP} respectively. Linearization definitions ({\tt \small lin}) indicate what strings are assigned to each of the meanings defined in the abstract syntax. 
To reduce same string declaration, 
the operator ({\tt \small oper}) section defines some placeholders for strings that can be used in linearization. The {\tt \small mkNP}, {\tt \small mkN}, {\tt \small mkV2}, etc. are standard constructors from ConstructorsEng/Jpn library which returns an object of the type {\tt \small NP}, {\tt \small N} or {\tt \small V2} respectively.

GF has been used in a variety of applications, such as query-answering systems, voice communication, language learning, text analysis and translation, natural language generation \cite{ranta2011grammatical,BurdenH11}, automatic translation\footnote{
   The MOLTO project: \url{http://www.molto-project.eu}
}. 

The translation from English to Italian can be performed as follows in the GF  API:


\begin{lstlisting}[escapechar=\#]
parse -lang=PeopleEng "Bill plays soccer" | linearize -lang=PeopleIta
  Bill gioca calcio
\end{lstlisting}

\begin{figure}[h]
  \centering
  \includegraphics[width=\columnwidth]{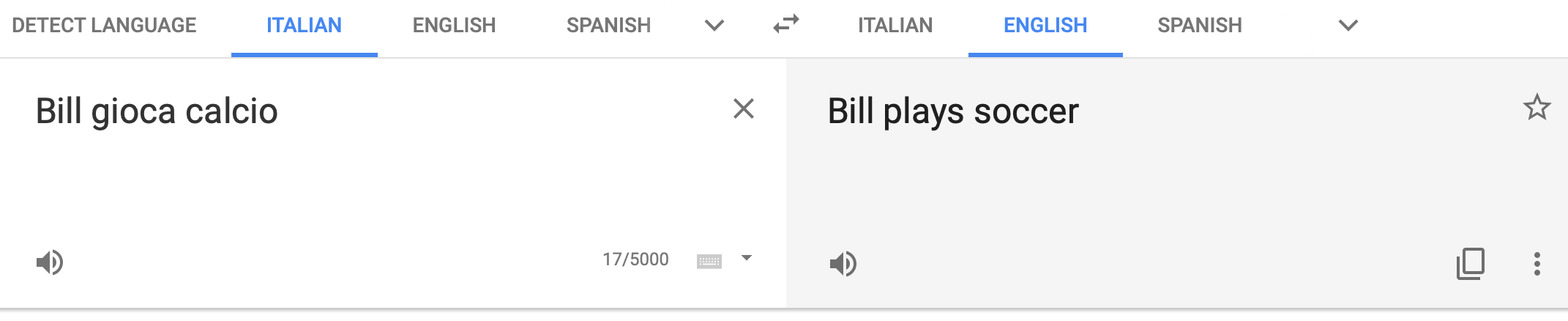}%
  \caption{Google translation for the Japanese sentence generated by GF. The two sentences in English and in Italian are the two representations of the meaning encoded in the abstract syntax.}
  \label{fig:translation}
\end{figure}

\noindent The above command line produces a syntax tree of the sentence \emph{``Bill plays soccer''}  then turn that tree into a {\tt \small PeopleIta} sentence (in Italian) which is displayed in the second line. 
Figure~\ref{fig:translation} shows the meaning in the abstract syntax is represented in Japanese and in Italian, i.e. the two strings represent the same meaning.

\section{Method}
\label{method}
To generate a sentence, we need a sentence structure and vocabularies. Our system is developed to emulate 
the process of a person learning a new language and has to make guesses to understand new sentences from time to time. For example, someone, who understands the sentence ``Bill plays a game'' would not fully understand the sentence ``Bill plays a popular board game'' without knowing the meaning of ``popular'' and ``board game'' but could infer that the latter sentence indicates that its subject plays a type of game. 

\begin{figure}[h]
  \centering
  \includegraphics[width=\columnwidth]{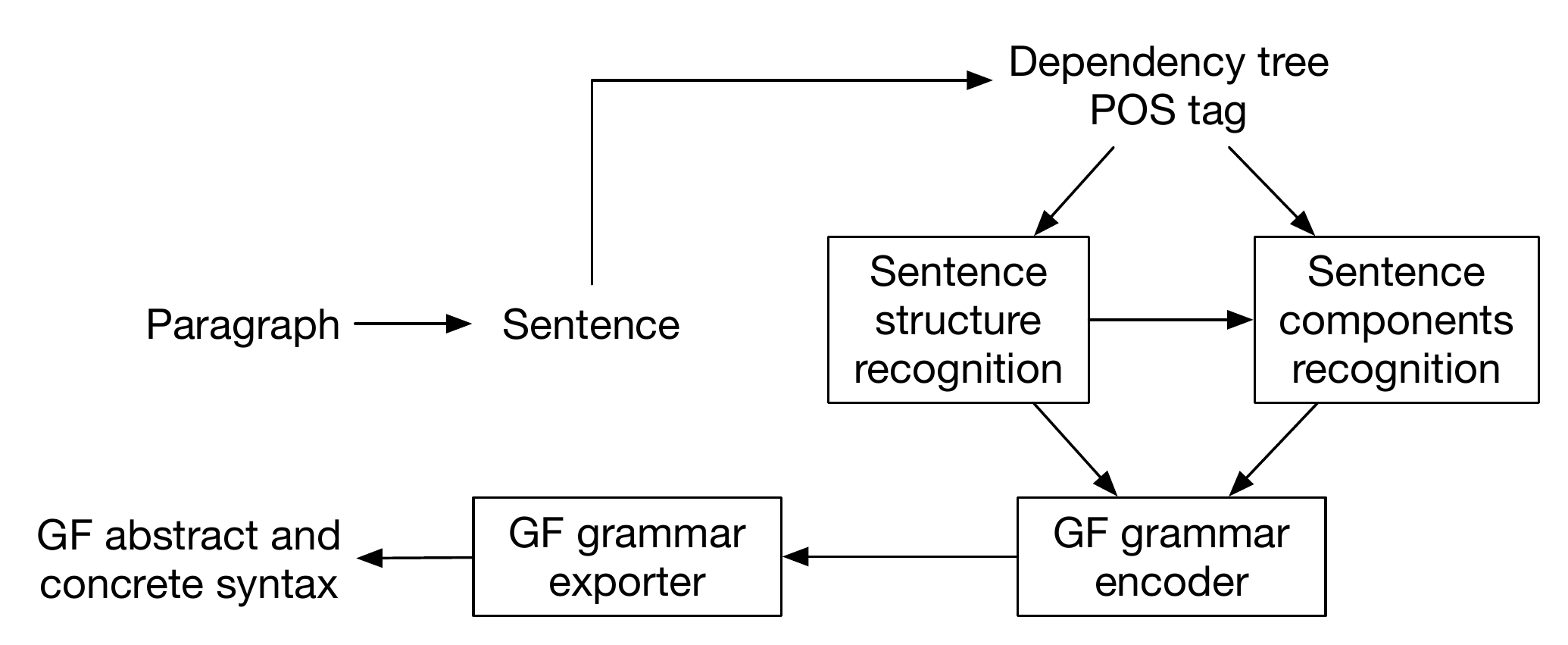}%
  \caption{System Overview}
  \label{fig:overview}
\end{figure}
The overall design of our system is given in Figure~\ref{fig:overview}. Given a paragraph, our system produces a GF program (a pair of an abstract and a  concrete syntax), which can be used for sentence generation. The system consists of two components,  understanding sentences and generating GF grammar. The first component is divided into two sub-components, one for recognizing the sentence structure and one for recognizing the sentence components. The second component consists of a GF grammar encoder and a GF grammar exporter. The encoder is responsible for generating a GF grammar for each sentence, while the exporter aggregates the grammars generated from the encoder, and produces a comprehensive grammar for the whole paragraph. 

\subsection{Sentence Structure Recognition}

The sentence structure recognition process involves 2 modules: natural language processing (NLP) module and logical reasoning on result from NLP module. In this paper, we make use of the Stanford Parser tools\footnote{\url{https://nlp.stanford.edu/software/lex-parser.shtml}} described in 
 \cite{chen2014fast,Nivre2016UniversalDV,schuster2016enhanced,de2006generating,Klein2002Fast} 

The NLP module tokenizes the input free text to produce a dependency-based parse tree \footnote{\url{https://en.wikipedia.org/wiki/Parse_tree }} and part-of-speech tag (POS tag). The dependency-based parse tree and the POS tag are then transform into an answer set program (ASP) \cite{GelfondL90} which contains only facts. Table \ref{tableTransformationNLPASP} shows the transformation of the result of NLP module into an ASP program for the sentence ``Bill plays a game''. In this table, \emph{nsubj, det, dobj} and \emph{punct} denote relations in the dependency-based parse tree, and mean \emph{nominal subject, determiner, direct object} and \emph{punctuation} respectively. Full description of all relations in a dependency-based parse tree can be found in the Universal Dependency website\footnote{\url{http://universaldependencies.org/}}. The second set of notations are the POS tag \emph{PRP, VBP, DT} and \emph{NN} corresponding to \emph{pronoun, verb, determiner} and \emph{noun}. Readers can find the full list of POS tag in Penn Treebank Project\footnote{\url{https://www.ling.upenn.edu/courses/Fall_2003/ling001/penn_treebank_pos.html}}.

{\small 
\begin{table}[h!]
\centering
  \begin{tabular}{|c|c|c|} 
  \hline
                  & \textbf{NLP result} & \textbf{ASP program} \\
  \hline
  \makecell{\textbf{Dependency} \\ \textbf{tree}}
  & \makecell{nsubj(plays-2,Bill-1) \\ ROOT(.-0,plays-2) \\ det(game-4,a-3) \\ dobj(plays-2,game-4) \\ punct(plays-2,.-5)} 
  & \makecell{nsubj(2,1). \\ det(4,3). \\ dobj(2,4). \\ punct(2,5).} \\
  
  \hline
  \textbf{POS tag}
  & \makecell{(Bill, PRP) \\ (plays, VBP) \\ (a, DT) \\ (game, NN) \\ (., .)} 
  & \makecell{pos\_tag(1,prp). \\ pos\_tag(2,vbp). \\ pos\_tag(3,dt). \\ pos\_tag(4,nn). \\ pos\_tag(5,punct).} \\
  \hline
  \end{tabular}
  \caption{Transformation from NLP result to asp program}
  \label{tableTransformationNLPASP}  
\end{table}
}

From the collection of the dependency atoms from the dependency-based parse tree, we determine the structure of a sentence using an ASP program, called $\Pi_1$ (Listing~\ref{list1}).  
\begin{lstlisting}[language=clingo,caption=Program $\Pi_1$,label=list1]
structure(1,1) :- nsubj(V,S).
structure(2,2) :- nsubj(V,S), dobj(V,O).
structure(3,3) :- nsubj(V1,S), xcomp(V1,V2), dobj(V2,O).
structure(4,2) :- nsubj(O,S), cop(O,TOBE).
structure(5,2) :- nsubjpass(V,S), auxpass(V,TOBE).
\end{lstlisting}
Each of the rule above can be read as \emph{if the right-hand side is true then the left-hand side must be true}.
These rules define five possible structures\footnote{
   These are the types of structures that we have implemented in our prototype. 
   Adding additional types will allow us to generate more complicated sentences. 
   This is left for our next work.  
} of a sentence represented by the atom \textit{structure(x,y)}. $x$ and $y$ in the atom \textit{structure(x,y)} denote the type of the structure and the number of dependency relations applied to activate the rule generating this atom, respectively. We refer to $y$ as the $i$-value of the structure. For example, $structure(1,1)$ will be recognized if the \emph{nsubj} relation is in the dependency-based parse tree; $structure(3,3)$ needs 3 dependency relations to be actived: \emph{nsubj}, \emph{xcomp} and \emph{dobj}. 
We often use structure \#$x$ to indicate a structure of type $x$. 

Together with the collection of the atoms encoding the relations in the dependency-based parse tree, $\Pi_1$ generates several atoms of the form $structure(x,y)$ for a sentence. Among all these atoms, an atom with the highest $i$-value represents the structure constructed using the highest number of dependency relations. And hence, that structure is the most informative structure that is recoginized for the sentence. 
Observe that $structure(1,1)$ is the most simplified structure of any sentence. 

\subsection{Sentence Components Recognition}
\label{subsectionComponentRecog}
The goal of this step is to identify the relationship between elements of a sentence structure and chunks of words in a sentence from the POS tags and the dependency-based parse tree. For example, the sentence ``Bill plays a game'' is encoded by a structure \#2 and we expect that  \textit{Bill},  \textit{plays}, and \textit{game} correspond to the subject, verb, and object, respectively.

We begin with recognizing the main words (components) that play the most important roles in the sentence based on a given sentence structure. This is achieved by program $\Pi_2$ (Listing~\ref{list2}). The first four rules of $\Pi_2$ determine the main subject and verb of the sentence whose structure is \#1, \#2, \#3, or \#5. Structure \#4 requires a special treatment since the components following \textit{tobe} can be of different forms. For instance, in ``Cathy is gorgeous,'' the part after \textit{tobe} is an adjective, but in  ``Cathy is a beautiful girl,'' the part after \textit{tobe} is a noun, though, with adjective \textit{beautiful}. This is done using the four last rules of $\Pi_2$. 

\begin{lstlisting}[language=clingo,caption=Program $\Pi_2$,label=list2]
2 { sub(S); verb(V) }         :- nsubj(V,S).
3 { sub(S); obj(O); verb(V) } :- nsubj(V,S), dobj(V,O).
2 { sub(S); verb(V) }         :- nsubjpass(V,S), auxpass(V,TOBE).  
4 { sub(S); obj(O); verb_1(V1); verb_2(V2) } 
                         :- nsubj(V1,S), xcomp(V1,V2), dobj(V2,O).
2 { sub(S); adj(O) }          :- nsubj(O,S), pos_tag(O,jj).
2 { sub(S); obj(O) }          :- nsubj(O,S), pos_tag(O,nn).
2 { sub(S); obj(O) }          :- nsubj(O,S), pos_tag(O,nns).
2 { sub(S); obj(O) }          :- nsubj(O,S), pos_tag(O,cd).
\end{lstlisting}

The result of program $\Pi_2$ is an one-to-one mapping of some of the words in the sentence into the importaint components of a sentence, called main components, i.e. subject, object and verb. The mapping is constructed by using the core arguments in Universal Dependency Relations \footnote{\url{https://universaldependencies.org/u/dep/}}. Since not every word in the sentence is in a core argument relation, there are some words in the sentence that are not in the domain of the mapping that $\Pi_2$ produces. We denote these words are complement components. To identify these words, we encode the Non-core dependents and Nominal dependents from Universal Dependency Relations into the set of rules in program $\Pi_3$.

Program $\Pi_3$ (Listing~\ref{list3}), together with the atoms extracted from the dependency-based parse tree such as $compound(P,N)$ ($N$ is compound noun at the position $P$ in the sentence), $amod(P,J)$ ($J$ is an adjective modifier), etc., is used to identify the complement components of the main components computed by $\Pi_2$ while maintaining the structure of the sentence created by $\Pi_1$.  For example, a complement of a noun could be another noun (as ``board'' in ``board game''), or an adjective (as ``popular'' in ``popular board game''), or a preposition (as ``for adults'' in ``board game for adults''). 

\begin{lstlisting}[language=clingo,caption=Program $\Pi_3$,label=list3]
noun_compound(N)        :- compound(pos,N).
adj_mod(JJ)             :- amod(pos,JJ).
noun_conjunction(N)     :- conj(pos,N).                        
preposition(COMP,IN)    :- nmod(pos,COMP), case(COMP,IN).
adverbial_modifier(ADV) :- advmod(pos,ADV).
\end{lstlisting}
The input of Program $\Pi_3$ is the position ($pos$) of the word in the sentence. Program $\Pi_3$ is called whenever there is a new complement component discovered. That way of recursive calls is to identify the maximal chunk of the words that support the main components of the sentence. The result of this module is a list of vocabularies for the next steps. 
 
\subsection{GF Grammar Encoder}
 
The goal of the encoder is to  identify appropriate GF rules for the construction of a GF grammar of a sentence given its structure and its components identified in the previous two modules. This is necessary since a sentence can be encoded in GF by more than one set of rules; for example, the sentence ``Bill wants to play a game'' can be encoded by the rules  \\
\centerline{
\textit{Bill} \ar NP, \textit{want} \ar VV, \textit{play} \ar V2, \textit{game} \ar NP
} 
and one of the sets of GF rules in the table below: 

{\small 
\begin{center} 
\begin{tabular}{|l|l|}\hline 
 V2 \ar NP \ar VP & V2 \ar NP \ar VP \\
 NP \ar VV \ar VP \ar Cl & VV \ar VP \ar VP \\
  & NP \ar VP \ar Cl \\
  \hline
\end{tabular} 
\end{center}
}

\noindent
In GF, NP, VV, V2, VP, and Cl stand for \emph{noun phrase}, \emph{verb-phrase-complement verb}, \emph{two-place verb}, \emph{verb phrase} and \emph{clause}, respectively. Note that although the set of GF grammatical rules can be used to construct a constituency-based parse tree \footnote{Constituency parsing aims to extract a constituency-based parse tree from a sentence that represents its syntactic structure \url{http://nlpprogress.com/english/constituency_parsing.html}}, the reverse direction is not always true. To the best of our knowledge, there exists no algorithm for converting a constituency-based parse tree to a set GF grammar rules. We therefore need to identify the GF rules for each sentence structure.

In our system, a GF rule is assigned to a structure initially (Table \ref{tableFirstLevelGFRule}). Each rule in Table~\ref{tableFirstLevelGFRule} represents the first level of the constituency-based parse tree. It acts as the coordinator for all other succeeding rules. 

\begin{table}
{\small 
  \begin{center} 
  \begin{tabular}{|c|l|} 
  \hline
  \textbf{Structure}    & \makecell[c]{\textbf{GF rules}} \\
  \hline
  \makecell{\#1} & \makecell[l]{NP \ar VP \ar Cl} \\
  \hline
  \makecell{\#2} & \makecell[l]{NP \ar V2 \ar NP \ar Cl} \\
  \hline
  \makecell{\#3} & \makecell[l]{NP \ar VV \ar V2 \ar NP \ar Cl} \\
  \hline
  \makecell{\#4} & 
  \makecell[l]{
    NP \ar AP \ar Cl and 
    NP \ar NP \ar Cl 
  } \\
  \hline
  \makecell{\#5} & \makecell[l]{NP \ar passiveVP \ar Cl} \\
  \hline
  \end{tabular}
  \end{center}
  }
  \caption{GF Rules Assigned to Each Structure}
  \label{tableFirstLevelGFRule}  
\end{table}

Given the seed components identified in Section \ref{subsectionComponentRecog} and the above GF rules, a GF grammar for each sentence can be constructed. However, this grammar can only be used to generate fairly simple sentences. For example, for the sentence ``Bill plays a popular board game with his close friends.'', a GF grammar for structure \#2 can be constructed, which can only generate the sentence ``Bill plays game.'' 
because it does not contain any complement components identified in Section~\ref{subsectionComponentRecog}. Therefore, we assgin a set of GF rules for the construction of each parameter in the GF rules in Table~\ref{tableFirstLevelGFRule}. The set of GF rules has to follow two conventions. The first one is after applying the set of rules to some components of the sentence, the type of the production is one of the type in Table~\ref{tableFirstLevelGFRule}, e.g. $NP$, $VP$, $Cl$, $V2$, \dots. The second convention is that the GF encoder will select the rules as the order from top to bottom in Table~\ref{tableExtendedGFRule}. Note that the encoder always has information of what type of input and output for the rule it is looking for.

For instance, we have ``game'' is the object (main components), and we know that we have to construct ``game'' in the result GF grammar to be a NP (noun phrase). Program $\Pi_2$ identifies that there are two complement components for the word ``game'', which are ``board'' and ``popular'', a noun and an adjective respectively. The GF encoder then select the set of rules: N \ar N \ar CN and A \ar AP to create the common noun ``board game'' and the adjective phrase first. The next rule is AP \ar CN \ar CN. The last rule to be applied is CN \ar NP. The selection is easily decided since the input and the output of the rules are pre-determined, and there is no ambiguity in the selection process.

\begin{table} 
  {\small 
    \begin{center} 
    \begin{tabular}{|l|l|}\hline 
    \multicolumn{2}{|c|}{For noun components} \\ 
    \hline
          N \ar N \ar CN  & CN:  \emph{common noun} \\ 
          N \ar NP & NP: \emph{noun phrase} \\ 
          AP \ar CN \ar CN &  AP: \emph{adjectival phrase} \\ 
          CN \ar NP &  \\
          NP \ar Adv \ar NP &  Adv: \emph{verb-phrase-modifying adverb} \\ 
          NP \ar NP \ar ListNP&  \\ 
          NP \ar ListNP \ar ListNP & \\ 
          Conj \ar ListNP \ar NP &  Conj: \emph{conjunction} \\ 
    \hline \hline
    \multicolumn{2}{|c|}{For verb components} \\ 
    \hline 
          VP \ar Adv \ar VP &  \\ 
      \hline \hline
    \multicolumn{2}{|c|}{For adjective components} \\ 
    \hline 
      A \ar AP & A: \emph{adjective} \\
      AdA \ar AP \ar AP  &  AdA: \emph{adjective-modifying adverb} \\ 
      \hline 
    \end{tabular} 
    \end{center}
  }
  \caption{Extended GF Rules}
  \label{tableExtendedGFRule}  
\end{table}  

The encoder uses the GF rules and the components identified by the previous subsections to produce different constructors for different components of a sentence. A part of the output of the GF encoder for the object ``game''  is
\begin{lstlisting}[language=gf]
Game = mkNP (mkNP popular_board_game_CN ) (ConstructorsEng.mkAdv 
       with_Prep (mkNP close_friend_CN )) ;
\end{lstlisting}
The encoder will also create the operators that will be included in the {\tt \small{oper}} section of
the GF grammar for supporting the new constructor. For example, the following operators will be generated for serving the {\tt \small Game} constructor above:
\begin{lstlisting}[language=gf]
  popular_A = mkA "popular" ;
  popular_AP = mkAP popular_A ;
  popular_board_game_CN = mkCN popular_AP board_game_N ;
  board_game_N = mkN "board game" "board games" ;
  close_A = mkA "close" ;
  close_AP = mkAP close_A ;
  close_friend_CN = mkCN close_AP friend_N ;
  friend_N = mkN "friend" "friends" ;
\end{lstlisting}

\subsection{GF Grammar Exporter}

The GF Grammar Exporter has the simplest job among all modules in the system. It creates a GF program for a paragraph using the GF grammars created for the sentences of the paragraph. By taking the union of all respective elements of each grammar for  each sentence, i.e., categories, functions, linearizations and operators, the Grammar Exporter will group them into the set of categories (respectively, categories, functions, linearizations, operators) of the final grammar.

\section{Experiments}
\label{usecase}

We describe our method of generating natural language in two applications. The first application is to generate a natural language description for workflow created by the system built in the Phylotastic project described in \cite{van2019padl}. Instead of requiring that the ontologies are annotated using Attempto, we use natural language sentences to annotate the ontologies. To test the feasibility of the approach, we also conduct another use case with the second ontology, that is entirely different from the ontologies used in the Phylotastic project. The ontology\footnote{Bookmarked URIs in Protege 5.5.0 Build beta-9 or \url{http://owl.man.ac.uk/2006/07/sssw/people}} is about people and includes descriptions for certain class.  

The second application targets the challenge of creating an abstract Wikipedia from the BlueSky session of 2018 International Semantic Web Conference \cite{Vrandecic18}. We create an intermediate representation that can be used to translate the original article in English to another language. In this use case, we translate the intermediate representation back to English and measure how the translated version stacks up again the original one. We assess the generation quality automatically with BLEU-3 and ROUGE-L (F measure). BLEU \cite{Papineni02bleu:a} and ROUGE \cite{lin2004rouge} algorithms are chosen to evaluate our generator since the central idea of both metrixes is ``the closer a machine translation is to a professional human translation, the better it is'', thus, they are well-aligned with our use cases' purpose. In short, the higher BLUE and ROUGE score are, the more similar the hypothesis text and the reference text is. In our use case, the hypothesis for BLEU and ROUGE is the generated English content from the intermediate representation, and the reference text is the original text from Wikipedia.

\subsection{NLG for Annotated Ontologies}
As described in \cite{van2019padl}, the author's system retrieves a set of atoms from an ASP program such as those in Listing~\ref{list6} where \emph{phylotastic FindScientificNamesFromWeb GET} was shortened to  \emph{service}, propagates the atoms, and constructs a set of sentences having similar structure to the sentence \emph{``The input of phylotastic FindScientificNamesFromWeb GET is a web link. Its outputs are a set of species names and a set of scientific names''}. In this sentence, \emph{phylotastic FindScientificNamesFromWeb GET} is the name of the service involved in the workflow of the Phylotastic project. All of the arguments of the atoms above are the names of classes and instances from Phylotastic ontology. 

\begin{lstlisting}[language=clingo,caption=Sample Set of Atoms,label=list6]
input(service, web_link).          typeof(web_link, url).
output(service, species_names).    typeof(species_names, names).
output(service, scientific_names). typeof(scientific_names, names).
\end{lstlisting}

We replace the original Attempto annotations with the natural language annotations as in 
Table \ref{table:annotation} and test with our system. 

With the same set of atoms as in Listing~\ref{list6}, our system generates the following description \emph{``Input of  phylotastic FindScientificNamesFromWeb GET is web link. Type of web link is url. Output of  phylotastic FindScientificNamesFromWeb GET is scientific names. Output of  phylotastic FindScientificNamesFromWeb GET is species names. Type of scientific names is names. Type of species name is names.''}.

\begin{table} {
  \small
  \begin{center} 
  
  \begin{tabular}{|c|c|} 
    \hline
    \makecell[c]{\textbf{Atom}} & \makecell[c]{\textbf{Annotation}} \\
    \hline
    \makecell[c]{input(service, \\
    web\_link).}
    & The input of \emph{service} is a web link \\
    \hline
    \makecell[c]{output(service, \\
    species\_names).}
    & The output of \emph{service} is species names \\
    \hline
    \makecell[c]{typeof(web\_link,  
    url).}
    & The type of web link is url \\
    \hline
    \end{tabular} 
    \end{center}
  }
  \captionsetup{justification=centering}
  \caption{Atoms from Phylotastic project and its annotation}
  \label{table:annotation}
\end{table}
 
 We also test our system with the people ontology as noted above. We extract all comments about people and replace compound sentences with simple sentences, e.g.,  \emph{``Mick is male and drives a white van''} is replaced by the two sentences   \emph{``Mick is male''} and \emph{``Mick drives a white van.''} to create a collection of sample sentences. We then use our system to generate a GF program which is used to generate sentences for RDF tuples. Sample outputs for some tuples are in Table \ref{table:people}. This shows that for targeted applications, our system could do a reasonable job.
 
\begin{table} {
  \small
  \begin{center} 
    \begin{tabular}{|c|c|} 
    \hline
    \makecell[c]{\textbf{Tuple}}                 & \makecell[c]{\textbf{Text}} \\
    \hline
    \makecell[c]{$($Kevin,has\_pet,Flossie$)$}   & Kevin has\_pets Flossie.  \\
    \hline
    \makecell[c]{$($Flossie,rdf$:$type,cow$)$}    & Flossie is cow. \\
    \hline
    \makecell[c]{$($Mick,reads,Daily\_Mirror$)$} & Mick reads Daily Mirror. \\
    \hline
    \end{tabular} 
  \end{center}}
  \captionsetup{justification=centering}
  \caption{Sample outputs for the people ontology.}
  \label{table:people}
\end{table}

\subsection{Intermediate Representation for Wiki Pages}

Since our system creates a GF program for a set of sentences, it could be used as an intermediate representation of a paragraph. This intermediate representation could be used by GF for automatic translation as GF is well-suited for cross-languages translation. On the other hand, we need to assess whether the intermediate representation is meaningful. This use case aims at checking the adequacy of the representation. To do so, we generate the English sentences from the GF program and evaluate the quality of these sentences against the original ones. We randomly select 5 articles from 3 Wikipedia portals: People, Mathematics and Food \& Drink. 

With the small set of rules introducing in this paper to recognize sentence structure, there would be very limited 4-gram in the generated text appearing in original Wikipedia corpus. Therefore, we use BLEU-3 with equal weight distribution instead of BLEU-4 to assess the generated content. Table \ref{tableBleuAssessableSentences} shows the summary of the number of assessable sentences from our system. Out of 62 sentences from 3 portals, the system cannot determine the structure 2 sentences in Mathematics due to their complexity. This low number of failure shows that our 5 proposed sentence structures effectively act as a lower bound on sentence recognition module. 

\begin{table}
{\small
  \begin{center}  
  \begin{tabular}{|c|c|c|c|} 
  \hline
                            & 
  \makecell{People}         & 
  \makecell{Mathematics}    & 
  \makecell{Food \& drink}  \\
  \hline
  \textbf{\#sentences}                                   & 15 & 24	& 23  \\
  \hline
  \makecell{\textbf{\#sentences} \\ \textbf{recognized}} & 15 & 22	& 23  \\
  \hline
  \makecell{\textbf{BLEU} \\ \textbf{assessable}}        & 10 & 15	& 11  \\
  \hline
  \end{tabular}
  \end{center}
}  
  \caption{BLEU assessable sentences }
  \label{tableBleuAssessableSentences}  
\end{table}

In terms of quality, Table \ref{tableScore} shows the average of BLEU and ROUGE score for each portal. Note that the average BLUE score is calculated only on BLEU assessable sentences, while average ROUGE score is calculated on the sentences whose structure can be recognized and encoded by our system. 
We note that the BLEU or ROUGE score might not be sufficiently high for a good quality translation. We believe that two reasons contribute to this low score. First, the present system uses fairly simple sentence structures. Second, it does not consider the use of relative clauses to enrich the sentences. This feature will be added to the next version of the system. 

\begin{table}
{\small
  \begin{center}  
  \begin{tabular}{|c|c|c|c|} 
  \hline
                            & 
  \makecell{People}         & 
  \makecell{Mathematics}    & 
  \makecell{Food \& drink}  \\
  \hline
  \textbf{BLEU}                   & 39.1 & 33.4	  & 52.2  \\
  \hline
  \makecell[c]{\textbf{ROUGE-1}}  & 20	 & 17.9	  & 14.6  \\
  \hline
  \makecell[c]{\textbf{ROUGE-2}}  & 6.7	 & 6.7	  & 5.5	  \\
  \hline
  \makecell[c]{\textbf{ROUGE-L}}  & 11.4 & 10.5   & 8.13  \\
  \hline
  \end{tabular}
\end{center}    
}
  \caption{BLUE and ROUGE score}
  \label{tableScore}
\end{table}

Table~\ref{tableCompare} summarizes the result of this use case. On the left are the paragraphs extracted from the Wikipedia page about Rice\footnote{\url{https://en.wikipedia.org/wiki/Rice}} in Food \& Drink, Decimal\footnote{\url{https://en.wikipedia.org/wiki/Decimal}} in Mathematics,  and about Alieu Ebrima Cham Joof\footnote{\url{https://en.wikipedia.org/wiki/Alieu_Ebrima_Cham_Joof}} from People. As we can see, the main points of the paragraphs are maintained.  

\begin{center}
  \begin{table}{
    \small
    \begin{center}
    \begin{tabular}{|p{9cm}|p{5.5cm}|}\hline
    \multicolumn{2}{|c|}{Rice} \\
    \hline 
          Rice is the seed of the grass species Oryza sativa (Asian rice) or Oryza glaberrima (African rice). & Rice is seed of grass species Oryza sativa \\
          \hline
          As a cereal grain, it is the most widely consumed staple food for a large part of the world's human population, especially in Asia. & it is widely consumed staple food for large part of human population of world in Asia \\
          \hline
          It is the agricultural commodity with the third-highest worldwide production (rice, 741.5 million tonnes in 2014), after sugarcane (1.9 billion tonnes) and maize (1.0 billion tonnes). & It is agricultural commodity with third-highest worldwide production after sugarcane \\
    \hline \hline
    \multicolumn{2}{|c|}{Decimal} \\
    \hline 
          The decimal numeral system is the standard system for denoting integer and non-integer numbers. & decimal numeral system is standard system.\\
          \hline
          It is the extension to non-integer numbers of the Hindu Arabic numeral system. & It is extension to non-integer number of Hindu-Arabic numeral system. \\
          \hline
          The way of denoting numbers in the decimal system is often referred to as decimal notation. & way is referred to decimal notation. \\
      \hline\hline
    \multicolumn{2}{|c|}{Alieu Ebrima Cham Joof} \\
    \hline
      Alieu Ebrima Cham Joof (22 October 1924 \u2013 2 April 2011) commonly known as Cham Joof or Alhaji Cham Joof, (pen name: Alh. A.E. Cham Joof) was a Gambian historian, politician, author, trade unionist, broadcaster, radio programme director, scout master, Pan-Africanist, lecturer, columnist, activist and an African nationalist. & Cham Joof is politician, author, unionist, broadcaster, radio programme director, scout master, Pan-Africanist, lecturer, columnist, activist, African nationalist and Gambian historian. \\
      \hline
       He advocated for the Gambia's independence during the colonial era. & He advocates for independence of Gambia during colonial era. \\
      \hline
    \end{tabular}
    \end{center}
  \caption{Original sentences extracted from Wikipedia and corresponding generated sentences}
  \label{tableCompare}
}
\end{table}
\end{center}

\section{Related Works}
\label{relatedwork} 

The systems developed in \cite{costa2018automatic,liu2018table,liu2019table} use statistical generation method to produce descriptions of tables or   explanation and recommendation from users' reviews of an item. All three systems are capable of generating high quality descriptions and/or explanations. In comparing to these systems, our system does not use the statistical generation method. Instead, we use Grammatical Framework for the generation task. A key difference between these systems and our system lies in the requirement of a large corpus of text in a specific domain for training and generation of these systems. Our system can work with very limited data and a wide range of domains.

Another method for generating natural language explanation for an question-answering system is proposed in \cite{DBLP:journals/corr/abs-1902-05715,cyras2019argumentation}. \cite{DBLP:journals/corr/abs-1902-05715} (\cite{DBLP:journals/corr/abs-1902-05715}) describes a system that can give reasonable and supportive evidence to the answer to a question asked to an image, while \cite{cyras2019argumentation} (\cite{cyras2019argumentation}) generates explanations for scheduling problem using argumentation. \cite{wang2017logic} (\cite{wang2017logic}) use ASP to develop a system answering questions in the do-it-yourself domain. These papers use templates to generate answers. The generated GF program generated by our system, that is used for the NLG task, is automatically created from a provided input.

The sophisticated system presented by \cite{arindam2019declarative} translates both question and the given natural language text to logical representation, and uses logical reasoning to produce the answer. Our system is similar to their system in that both employ recent developments of NLP into solving NLG problems.

\section{Conclusions and Future Work}
\label{conclusion} 

We propose a system implemented using answer set programming (ASP) and Grammatical Framework (GF), for automatic generation of  natural language descriptions in applications targeting mainstream users. 
The system does not require a large corpus for the generation task and can be used in different types of applications. 

In the first type of applications, the system can work with annotated ontologies to translate a set of atoms---representing the answer to a query to the ontology---to a set of sentences. To do so, the system extracts the annotations related to the atoms in the answer and creates a GF program that is then used to generate natural language description of the given set of atoms. 
In the second type of applications, the system receives a paragraph of text and generates an intermediate representation---as a GF program---for the paragraph, which can be used for different purpose such as cross-translation, addressing a need identified in \cite{Vrandecic18} .  

Our use cases with different ontologies and Wikipedia portals provide encouraging results. They also point to possible improvements that we plan to introduce to the next version of the system. We will focus on processing relative clauses and enriching the set of sentence structures, especially for compound and complex sentences.

\bibliographystyle{eptcs}
\bibliography{van}
\end{document}